# A real-time warning system for rear-end collision based on random forest classifier*

**Fateme Teimouri, Mehdi Ghatee†**

Department of Computer Science, Amirkabir University of Technology, Tehran, Iran

**Abstract**

Rear-end collision warning system has a great role to enhance the driving safety. In this system some measures are used to estimate the dangers and the system warns drivers to be more cautious. The real-time processes should be executed in such system, to remain enough time and distance to avoid collision with the front vehicle. To this end, in this paper a new system is developed by using random forest classifier. To evaluate the performance of the proposed system, vehicles trajectory data of 100 car's database from Virginia tech transportation institute are used and the methods are compared based on their accuracy and their processing time. By using TOPSIS multi-criteria selection method, we show that the results of the implemented classifier is better than the results of different classifiers including Bayesian network, naive Bayes, MLP neural network, support vector machine, nearest neighbor, rule-based methods and decision tree. The presented experiments reveals that the random forest is an acceptable algorithm for the proposed driver assistant system with 88.4% accuracy for detecting warning situations and 94.7% for detecting safe situations.

**Keywords:** Rear-end collision; Driver assistant systems; Data mining; Classification algorithms; TOPSIS.

## 1. Introduction

Road safety is an important subject all over the world. The general parameters influential on road accidents are driver, vehicle, road and environment [1]. According to the recent research studies about the accidents' reasons, human fault effects in almost 93% of accidents and it is the main reason in 75% of accidents [2].Using the safety systems installed on vehicles, it is possible to decrease the rate of accidents. These safety systems are divided into two groups, passive safety systems and active safety systems. The former helps people stay alive and uninjured in a collision and the latter helps drivers avoid accidents [3].However it is already revealed that most of passive safety systems including seat belt and airbag are losing their capabilities. In addition it is impossible to improve their safety with reasonable expenses. So currently active systems like advanced driver assistance system

*This paper was partially supported by Intelligent Transportation System Research Institute, Amirkabir University of Technology, Tehran, Iran.

† Corresponding author: Mehdi Ghatee, Department of Computer Science, Amirkabir University of Technology, No 424, Hafez Ave, Tehran 15875-4413, Iran Fax: +98216497930, Email: ghatee@aut.ac.ir, URL: http://www.aut.ac.ir/ghatee

(ADAS) has been followed a lot [4]. The main purpose of this system is reducing driver's faults with provided packages or information about dangerous situations ahead. Collision warning system as a subsystem of ADAS helps drivers to evaluate non-obvious dangerous situations to reduce related human's faults.

According to the wide variety of car accidents and their results, it is necessary to apply dedicated warning systems for all of the accidents types. Based on NHTSA‡ report in 2012, about 32.9 % of car accidents (or 1.8 million) are rear-end collisions [5]. When a rear-end collision happens, the probability of next contentions increases which increases the difficulty. Due to the importance of rear-end collisions, in the current paper an intelligent and online ADAS for rear-end collision warning is proposed, see e.g., [6] to see the effects of such systems. However the process time of a rear-end collision warning is very important in its effects. A poorly timed warning may actually undermine driver safety [7]. Too soon or too frequent alarms (false alarms) which both bother driver or too late alarms (missed alarms) which make it dangerous for drivers, both make driver to avoid using these systems [8]. As a result finding a balance between opportune alarming (not too soon, not too late) and detecting dangerous situations which both mean an efficient alarming system, is important and is noticed in designing the proposed systems.

Following [9], [10], [11] and [12], we have classified the existing algorithms for rear-end collision warnings into two groups including perceptual-based and kinematic-based. In the former, there are some criteria to evaluate driver performance. For each criterion, there is a threshold value and when the value of criterion is less than that threshold, warning will be issued. Time-to-collision (TTC) [13] and time-gap (TG) [14] are two criteria for analyzing the performance of a driver who is following a preceding vehicle [15]. A variety of existing studies have attempted to identify rear-end collision situations using time-to-collision, time-gap or both of them. [16] and [17] calculated a warning distance between vehicles based on critical threshold for TTC and these algorithms are called respectively Honda algorithm and Hirst and Graham algorithm. [18] and [19] proposed a vision based forward collision warning which uses TTC and possible collision course to trigger a warning. [20] used TTC to judge collision threat and to estimate potential effectiveness of a forward collision avoidance system with a forward collision warning system, a brake assist and autonomous braking functionality. [21] presented a methodology to estimate rear-end crash probability based on an exponential decay function using TTC. [22] used a non-dimensional warning index and TTC for determining driving situations and the main idea of this system is to maintain a specified TG between vehicles. [23] developed a fuzzy rear-end collision warning system using TTC and TG in order to warn driver of a possible collision.

However the latter works based on kinematic information of two vehicles, taking pre-determined driver's reaction time and maximum deceleration rate to calculate safe distance. Safe distance which is calculated differently in each algorithm, is taken as critical threshold and when the distance between two vehicles is less than this threshold, the warning alarm will be issued. [24], [25] and [8] calculated a warning distance between vehicles based on different scenarios and these algorithms are called respectively stop distance algorithm, Mazda algorithm and PATH algorithm. A vision-based ADAS with forward collision warning function was developed in [26] applying headway distance estimation model to detect the potential forward collision. Issuing forward collision warning based on distance was also proposed in [27]. [28] used safety distance to calculate rear-end collision risk index, then fuzzy-clustering algorithm has been applied to identify the rear-end collision risks level.

‡National Highway Traffic Safety Administration

However, in the previous algorithms, the accuracy of detecting warning and safe situations is not acceptable because of using some constants and pre-defined values for parameters and threshold values. Although there are a lot of researches on rear-end collisions detection and alarming, there is no enough attention to extract the knowledge about warning and safe situations. Such knowledge can be extracted from accidents and vehicles trajectory data. Data mining algorithms can be used to extract patterns from vehicles trajectory data to define the threshold values for parameters and to find the criteria which are more matched with reality. This is the most important contribution of this paper to used data mining to enhance the rear-end collision warning system.

The structure of the paper is as follows. The proposed methodology for rear-end collision warning system is presented in the next section. Section 3 optimizes the classification of movement situations into warning and safe classes and describes evaluation of the proposed methodology. The final section is dedicated to a brief conclusion of the paper.

## 2. Rear-end collision warning system

The vehicles continuously interact with the other vehicles to realize car-following and lane-changing, etc. When these interactions are not stable, collision is possibly happened. Therefore it is possible to evaluate the potential of collision with analyzing the vehicle's motion and unsafe situations [21]. Data mining techniques are appropriate tools to detect warning situations based on the vehicles trajectory data. These methods are used to analyze a large volume of data and to extract patterns and rules which shape the knowledge [29]. Among data mining techniques, classification algorithms are the most famous which can be used for our purpose. In the proposed system by classification algorithms and cost sensitive learning methods, we can extract warning situations for rear-end collisions and safe situations. In our study, 100 car's data base is considered [30]. To classify the situations, we follow the steps given in Figure 1. Also the details of its modules will be discussed in the next subsections.

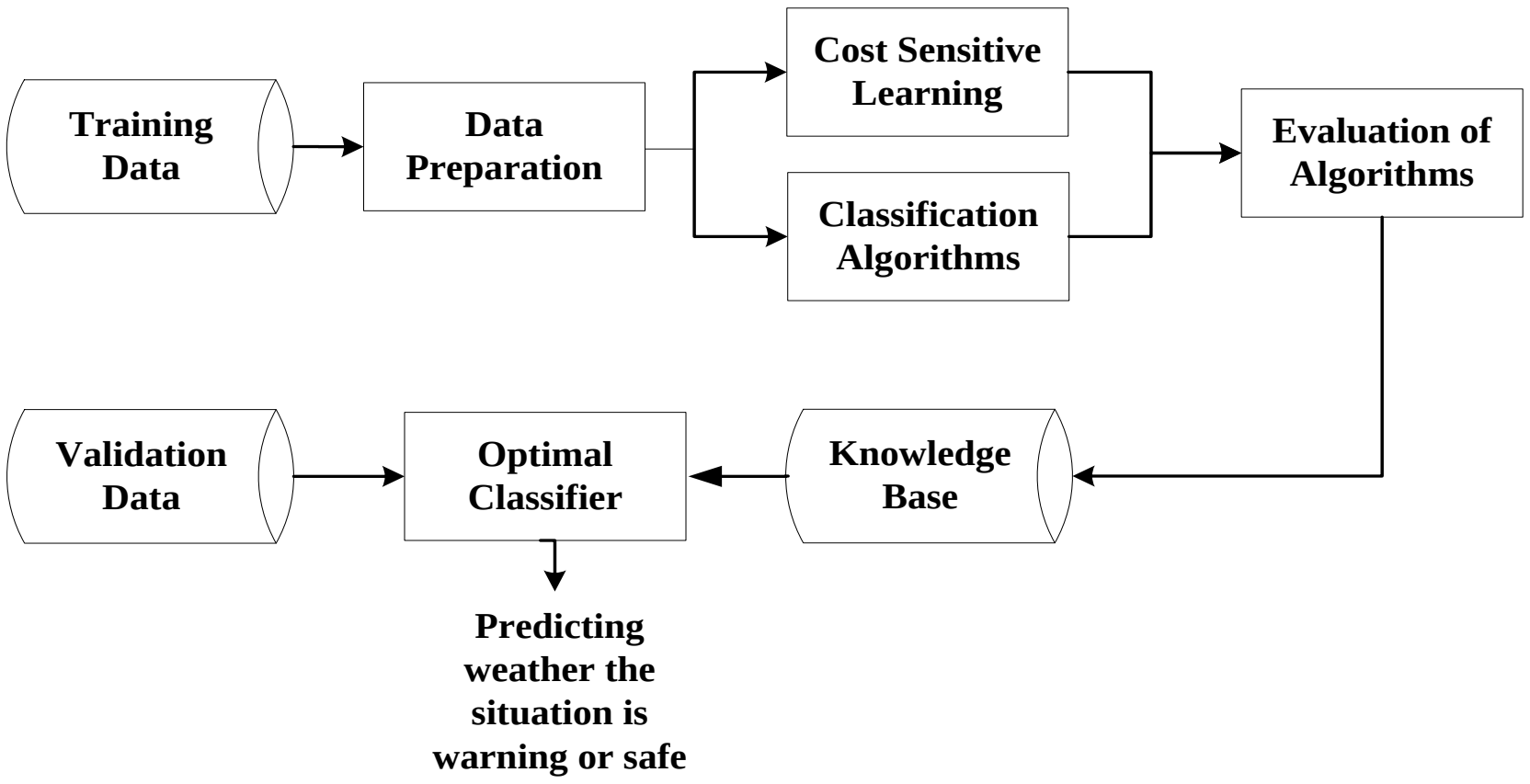

Figure 1 The structure of the proposed rear-end collision warning system

### 2.1. Cost-sensitive learning module

Usually cost-sensitive learning method is used on databases with imbalanced class distribution [31]. In cost-sensitive learning, the class with less elements is considered as positive class and the other with more elements is taken as negative class into account. Often misclassification of actual positive classes but predicted as negative one is greater than misclassification of actual negative classes but predicted as positive one [32]. The cost of classification can be calculated with a cost matrix denoted in Table 1 .

Table 1 Confusion/Cost matrix for a two-class problem

| | **Positive prediction** | **Negative prediction** |
|---|---|---|
| **Positive class** | True Positive (TP) | False Negative (FN) |
| **Negative class** | False Positive (FP) | True Negative (TN) |

In the database which is being used for classification, we used cost-sensitive learning because the amount of data for safe situations is 5 times of warning situations and the cost matrix is demonstrated in Table 2 .

Table 2 Confusion/Cost matrix for warning and safe situations problem

| | **Warning prediction** | **Safe prediction** |
|---|---|---|
| **Warning class** | 0 | 5 |
| **Safe class** | 1 | 0 |

## 2.2. Classification algorithms module

All classification methods can be used to find a relation between input and output properties. In the proposed rear-end collision warning system, we implement random forest classifier which is an ensemble classifier including several unpruned trees. Each tree is independently built based on bootstrap samples of training dataset and the best split at each node is calculated within a randomly selected subset of descriptors. After trees are grown, predictions for test data is made by the majority vote of ensemble tees. To know more about this algorithm, see [33], [34]. In the next section we show that this algorithm is preferable on other classifiers for the proposed warning system. For this aim, classification algorithms in Weka[§] are called. Details of some examined classification methods are given as follows:

- Bayesian network: It is a directed acyclic graph (DAG) which illustrates a factorization of a joint probability distribution (JPD). Given a sufficiently large dataset, the Bayesian network can learn by structural learning and parameter learning. More details can be found in [35], [36].
- Naïve Bayes: It is a simple probabilistic method which assumes class conditional independence. This classifier predicts the class with the highest posterior probability which is calculated by Bayes theorem. For more study, see [37].
- Multi-layer perceptron: It is a feed-forward neural network with one or multiple hidden layers including different number of neurons. The best algorithm for tuning the weights in this network is backpropagation algorithm. More details have been presented in [38], [39].
- Support Vector Machine: It uses sequential optimization algorithm to find an optimal hyper plane that correctly classifies data points by separating the points of two classes as much as possible. For more information, see [40].
- K-nearest neighbor: Among different methods of supervised learning, the nearest neighbor has the highest efficiency because there is no preliminary assumption about training data distribution. For details, see [41].
- JRip rule-based (RIPPER[**]): It is a rule-based method in which there is a set of rules and if a sample has the properties of one of the rules, that sample is a member of rule's class otherwise it would not be counted as a rule's class. The RIPPER algorithm is a two stages algorithm to reduce errors by iteratively pruning the design space and extracting the knowledge of rules. For more research, see [42].
- C4.5 decision tree: It constructs decision tree in a top-down manner with a divide and conquer strategy. The construction begins by evaluating each attribute using a criterion known as information gain ratio to determine best attribute for classifying the training samples. This attribute is chosen for the root node of the tree. Then decision tree splits the data into subsets according to value of chosen attribute, and the process repeats for each child. Details of C4.5 have been given in [43], [44].

---

[§] http://wiki.pentaho.com/

[**] Repeated Incremental Pruning to Produce Error Reduction

### 2.3. Evaluation measures

To evaluate the results of the different classifiers, the data set is divided into training data set and validation data. Accuracy criterion (Equation (1)), sensitivity criterion (Equation(2)), specificity criterion (Equation (3)) and also processing time of building classifier model and classifying validation data are used to compare the results in this paper.

$$Accuracy = \frac{TruePositive + TrueNegative}{TruePositive + TrueNegative + FalsePositive + FalseNegative} \quad (1)$$

$$\text{Sensitivity} = \frac{TruePositive}{TruePositive + FalseNegative} \quad (2)$$

$$\text{Specificity} = \frac{TrueNegative}{TrueNegative + FalsePositive} \quad (3)$$

Moreover to evaluate the different methods in the proposed system, the properties of data samples are used for data mining process. These properties include the follower vehicle's speed, the relative distance between two vehicles (follower and leader vehicles), the relative speed between two vehicles, time to collision and time gap. Time to collision (TCC) is the time interval between two vehicles colliding together without changing the movement direction and speed. TTC is calculated by Equation(4).

$$TTC(t) = \frac{\Delta x}{-\Delta v} = \frac{x_l - x_f}{v_f - v_l} \quad (4)$$

In addition, based on Equation (5), the time gap (TG) is the time interval between the follower and the leader vehicles without any change in the situations.

$$TG(t) = \frac{\Delta x}{v_f} = \frac{x_l - x_f}{v_f} \quad (5)$$

In the next section, we use these measures in simulation experiments.

## 3. Simulation results

In this section, for installing the proposed rear-end collision warning system, firstly the appropriate number of hidden layer’s neurons for neural network techniques and the appropriate K for K-nearest neighbor method are obtained and then we compare classifiers based on different measures. Based on this comparison, the reasonable classifier is selected to install the proposed rear-end collision warning system.

### 3.1. The experimental data

In this section, we have used 100 car's database from Virginia tech transportation institute [45] to examine the data of vehicles trajectory. This database includes the information of 68 crashes and 760 near-crashes. In these near-crashes, a fast movement such as braking or lane changing avoids an accident. To use crashes' data, it is possible to use reinforcement learning. In reinforcement learning, the behaviors aligned with final goals will be rewarded and the other behavior farther from final goal will be penalized. For the cases that the driver's reaction is lane changing, the other lane properties should be taken into account to analyze the risky behavior. However, such data are not given in 100 car's database. So in the experiment of the current paper, we use the near-crashes data as the rear-end collisions where driver's reaction is braking. After such preprocess on 100 car's database, we have obtained 39938 data samples. The properties of these samples are used for data mining process in the proposed system as mentioned in Subsection 2.3.

Moreover, to install supervised learning methods, the sample data are divided into two classes: warning class or safe class. Each near-crashes' data includes 30 seconds trajectory before event, event and 10 seconds trajectory after event. Event means the time interval between leader vehicle's starting to stop, therefore rear-end collision starts to happen, until when the follower's driver does some fast actions like braking to avoid colliding the leader vehicle. Obviously, before and after event, there is not necessary to warn, however the warning is necessary during the event. So, the warning level is considered as zero or one. Zero is assigned to all time interval 30 seconds before event and 10 seconds after event. One is also assigned to the time interval of event.

### 3.2. The selection methodology for classifiers

To tune up the classifiers and to compare the results of different classifiers, the data set is divided into training data set and validation data set to create and validate the models, respectively. In this study, to guarantee independency between the results and the data, three scenarios are taken into account in which the training data are selected randomly with 65%, 70% and 80% of the existence data. The remaining data is used as validation data. Then to select the best classifier for each scenario, based on the assigned weights to all criteria, it is possible to use TOPSIS technique [46], [47] which is a famous method for solving multi-criteria decision making problems. To assign weights to the different criteria, there are different cases. In this study, we study the following 4 assumptions on the weights of criteria:

- Assumption 1: All three criteria have the same weight and importance,
- Assumption 2: The specificity and sensitivity are important and have the same weights,
- Assumption 3: All three criteria are important but specificity has greater weight than sensitivity and sensitivity hast greater weight than processing time,
- Assumption 4: All three criteria are important but sensitivity has greater weight than specificity and specificity has greater weight than processing time.

### 3.3. Experiment results on multi-layer neural network

An artificial neural network with a single hidden layer can be used to approximate every non-linear function by a pre-determined precision degree [48], however the number of neurons in the corresponding layer is not easy to obtain. In this study we try to obtain efficient number of neurons in the single hidden layer in the proposed rear-end collision warning system. By trial and error with different numbers of neurons in hidden layer, we trained the system and evaluated their performance. In what follows, we consider a three layer perceptron neural network with different number of neurons in hidden layer: (5, 10, 15, 20, 25, 30, 35, 40, 45, 50, 55, 60, 65, 70, 75, 80) and for all of the experiments, we check three scenarios as mentioned in the previous subsection. Figure 2 shows the simulation results for the three layer perceptron. Obviously, the highest accuracy and specificity of the first scenario are happened for a neural network with 75 neurons in the hidden layer and the highest sensitivity is reported for a network with 60 neurons in hidden layer. For the second scenario, the highest accuracy and specificity are regarded to the 40 neurons in the hidden layer and highest sensitivity is for 35 neurons. In the third scenario, the highest accuracy and specificity are obtained with 15 neurons in the hidden layer and highest sensitivity is for 25 neurons. In addition, Figure 3 compares the processing time for three scenarios in which the lowest processing time happens for 5 neurons in the hidden layer which is a trivial result.

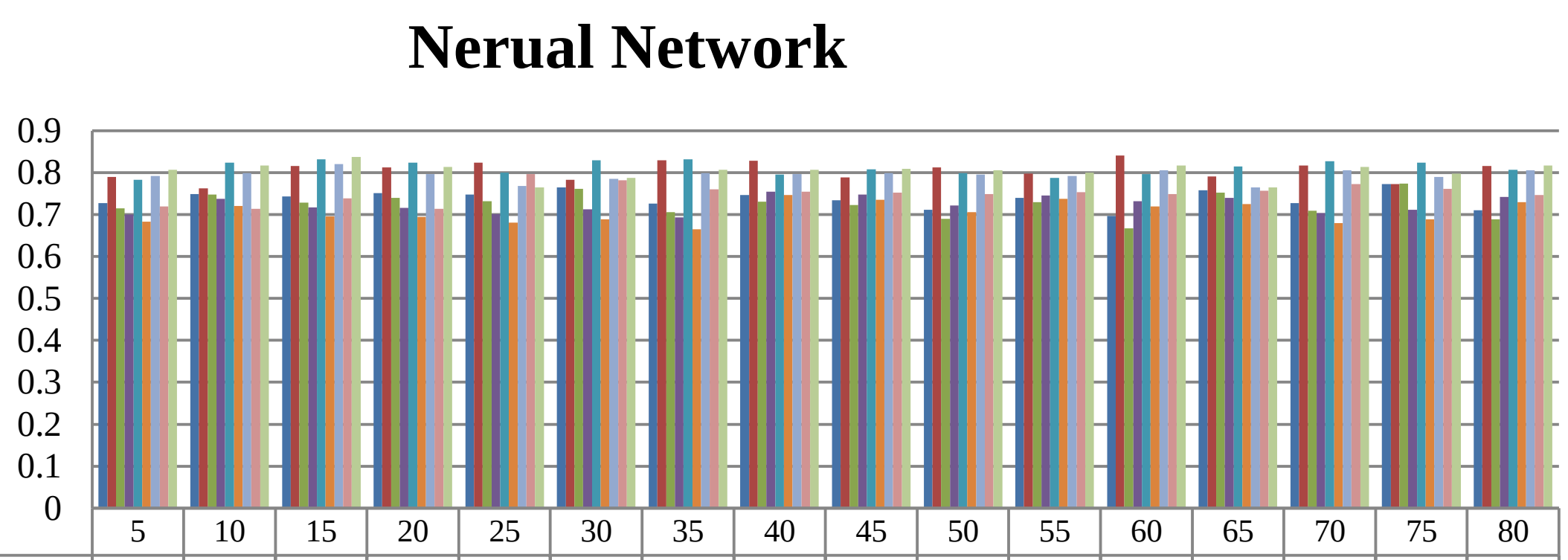

| | 5 | 10 | 15 | 20 | 25 | 30 | 35 | 40 | 45 | 50 | 55 | 60 | 65 | 70 | 75 | 80 |
|---|---|---|---|---|---|---|---|---|---|---|---|---|---|---|---|---|
| 1st Scenario: Accuracy | 0.727 | 0.749 | 0.743 | 0.751 | 0.747 | 0.764 | 0.726 | 0.746 | 0.734 | 0.711 | 0.74 | 0.696 | 0.758 | 0.727 | 0.773 | 0.71 |
| 1st Scenario: Sensitivity | 0.789 | 0.762 | 0.815 | 0.812 | 0.823 | 0.783 | 0.829 | 0.828 | 0.788 | 0.812 | 0.797 | 0.84 | 0.791 | 0.817 | 0.772 | 0.815 |
| 1st Scenario: Specificity | 0.715 | 0.747 | 0.728 | 0.739 | 0.732 | 0.761 | 0.706 | 0.73 | 0.723 | 0.69 | 0.729 | 0.667 | 0.752 | 0.709 | 0.774 | 0.689 |
| 2nd Scenario: Accuracy | 0.7 | 0.737 | 0.717 | 0.716 | 0.701 | 0.712 | 0.693 | 0.754 | 0.747 | 0.721 | 0.745 | 0.732 | 0.739 | 0.703 | 0.711 | 0.742 |
| 2nd Scenario: Sensitivity | 0.783 | 0.823 | 0.831 | 0.823 | 0.8 | 0.829 | 0.832 | 0.795 | 0.808 | 0.798 | 0.787 | 0.796 | 0.814 | 0.827 | 0.824 | 0.806 |
| 2nd Scenario: Specificity | 0.683 | 0.72 | 0.695 | 0.694 | 0.681 | 0.689 | 0.665 | 0.746 | 0.735 | 0.706 | 0.737 | 0.719 | 0.725 | 0.679 | 0.689 | 0.729 |
| 3rd Scenario: Accuracy | 0.792 | 0.799 | 0.82 | 0.796 | 0.768 | 0.785 | 0.798 | 0.796 | 0.799 | 0.795 | 0.792 | 0.805 | 0.764 | 0.805 | 0.79 | 0.805 |
| 3rd Scenario: Sensitivity | 0.719 | 0.714 | 0.738 | 0.714 | 0.796 | 0.781 | 0.76 | 0.754 | 0.752 | 0.749 | 0.753 | 0.749 | 0.757 | 0.773 | 0.761 | 0.746 |
| 3rd Scenario: Specificity | 0.807 | 0.817 | 0.837 | 0.813 | 0.764 | 0.787 | 0.807 | 0.806 | 0.809 | 0.805 | 0.8 | 0.817 | 0.765 | 0.813 | 0.797 | 0.817 |

Figure 2 Accuracy, sensitivity and specificity of three layer perceptron with different hidden neurons

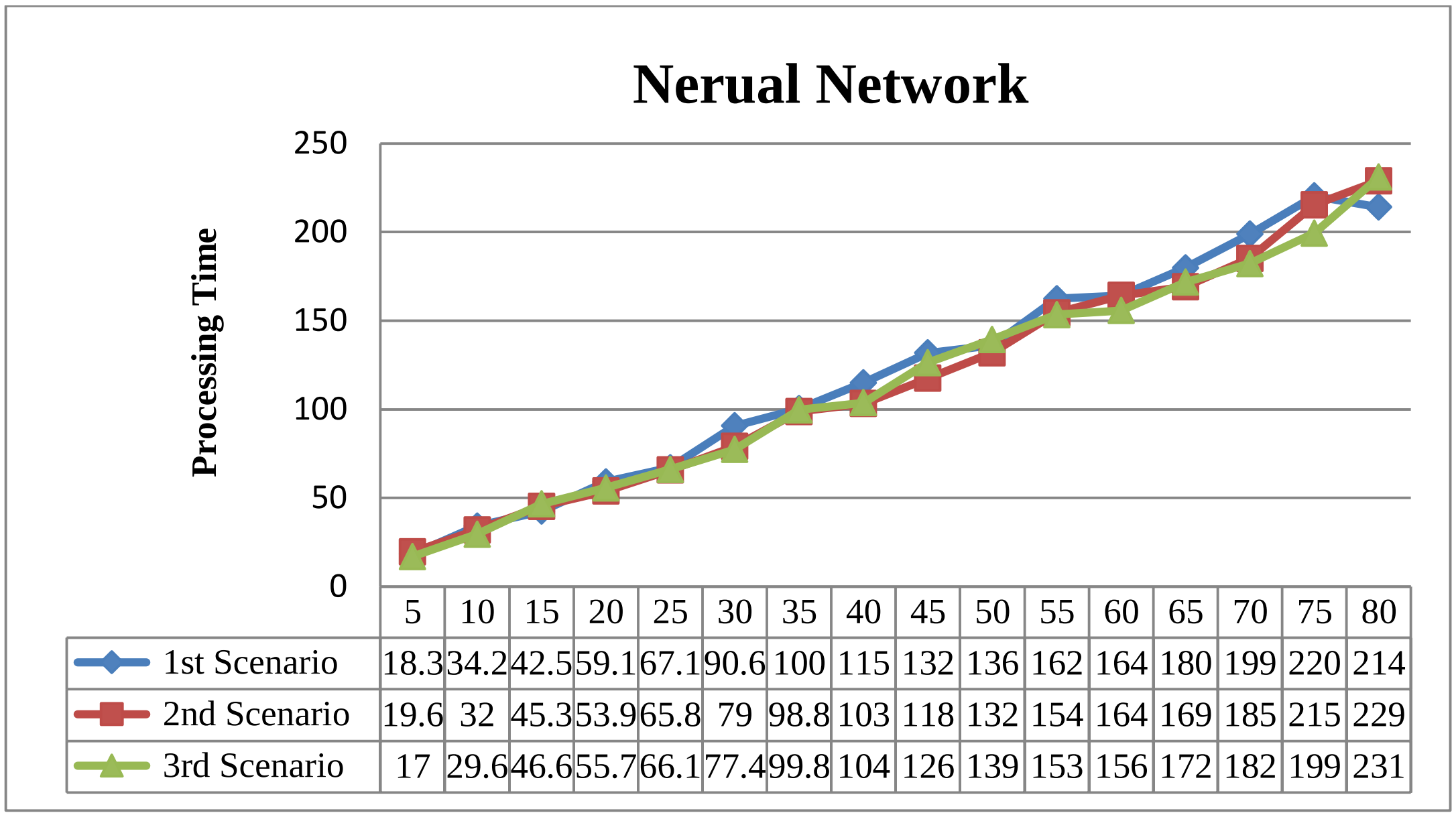

Figure 3 Processing Time of three layer perceptron with different hidden neurons

In the presented case, for each scenario, there is a decision matrix including 16 alternatives and 3 different criteria. Alternatives are the number of hidden neurons and the criteria are the sensitivity, the specificity and the processing time. Accuracy is not taken as a criterion into account because it is a combination of sensitivity and specificity. The decision matrix values can be extracted from Figure 2 and Figure 3. For all of the assumptions the weights of criteria are given in Table 3 .

Table 3 Weight matrix for different criteria for the considered assumptions and the best number of neurons in the hidden layer.

| **Assumptions Index** | **Sensitivity** | **Specificity** | **Processing Time** | **Third Scenario** | **Second Scenario** | **First Scenario** |
|---|---|---|---|---|---|---|
| 1 | $\frac{1}{3}$ | $\frac{1}{3}$ | $\frac{1}{3}$ | 5 neurons | 10 neurons | 5 neurons |
| 2 | $\frac{1}{2}$ | $\frac{1}{2}$ | 0 | 70 neurons | 45 neurons | 20 neurons |
| 3 | $\frac{2}{6}$ | $\frac{3}{6}$ | $\frac{1}{6}$ | 5 neurons | 10 neurons | 10 neurons |
| 4 | $\frac{3}{6}$ | $\frac{2}{6}$ | $\frac{1}{6}$ | 5 neurons | 10 neurons | 5 neurons |

Applying TOPSIS method for all of the mentioned assumptions shows that the best number of neurons of the hidden layer of the corresponding neural networks is different. SeeTable 3 including the best number of hidden neurons based on our experiments.

### 3.4. Experiment results on k-nearest neighbor method

In k-nearest neighbor method, similar to the way of tuning up the parameters in the last subsection, it is possible to find the best k to find the reasonable results. To this aim, we simulate this classifier for different $k \in \{1,2,3,4,5\}$ and for the previous three different scenarios. Then for each scenario, the best value of k has been obtained. The weights of criteria are similarly defined based on the assumptions given in Table 3 . Figure 4 shows the simulation results for the k-nearest neighbor. Obviously in all three scenarios, the greatest accuracy and specificity is obtained for k=1 and the maximum sensitivity is fetched for k=5. Thus as much as k increases, the accuracy and the specificity decreases and the sensitivity increases. In Figure 5 the processing times of all different k-nearest neighbor method for all of the scenarios and different k are illustrated. As one can note that there is not a great difference between these experiments in terms of processing times.

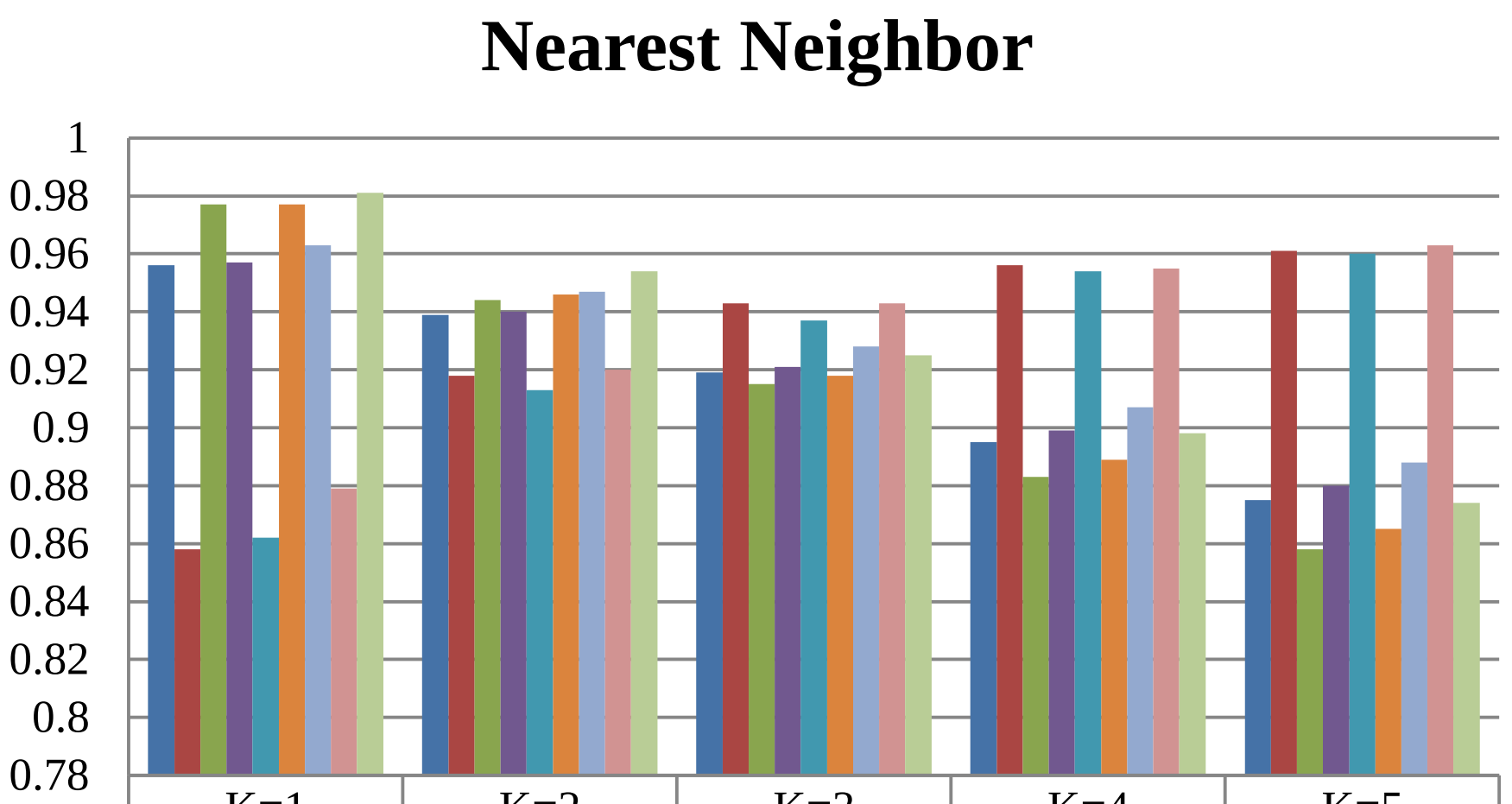

| | K=1 | K=2 | K=3 | K=4 | K=5 |
|---|---|---|---|---|---|
| 1st Scenario: Accuracy | 0.956 | 0.939 | 0.919 | 0.895 | 0.875 |
| 1st Scenario: Sensitivity | 0.858 | 0.918 | 0.943 | 0.956 | 0.961 |
| 1st Scenario: Specificity | 0.977 | 0.944 | 0.915 | 0.883 | 0.858 |
| 2nd Scenario: Accuracy | 0.957 | 0.94 | 0.921 | 0.899 | 0.88 |
| 2nd Scenario: Sensitivity | 0.862 | 0.913 | 0.937 | 0.954 | 0.96 |
| 2nd Scenario: Specificity | 0.977 | 0.946 | 0.918 | 0.889 | 0.865 |
| 3rd Scenario: Accuracy | 0.963 | 0.947 | 0.928 | 0.907 | 0.888 |
| 3rd Scenario: Sensitivity | 0.879 | 0.92 | 0.943 | 0.955 | 0.963 |
| 3rd Scenario: Specificity | 0.981 | 0.954 | 0.925 | 0.898 | 0.874 |

Figure 4 Accuracy, sensitivity and specificity of nearest neighbor classifier with different K

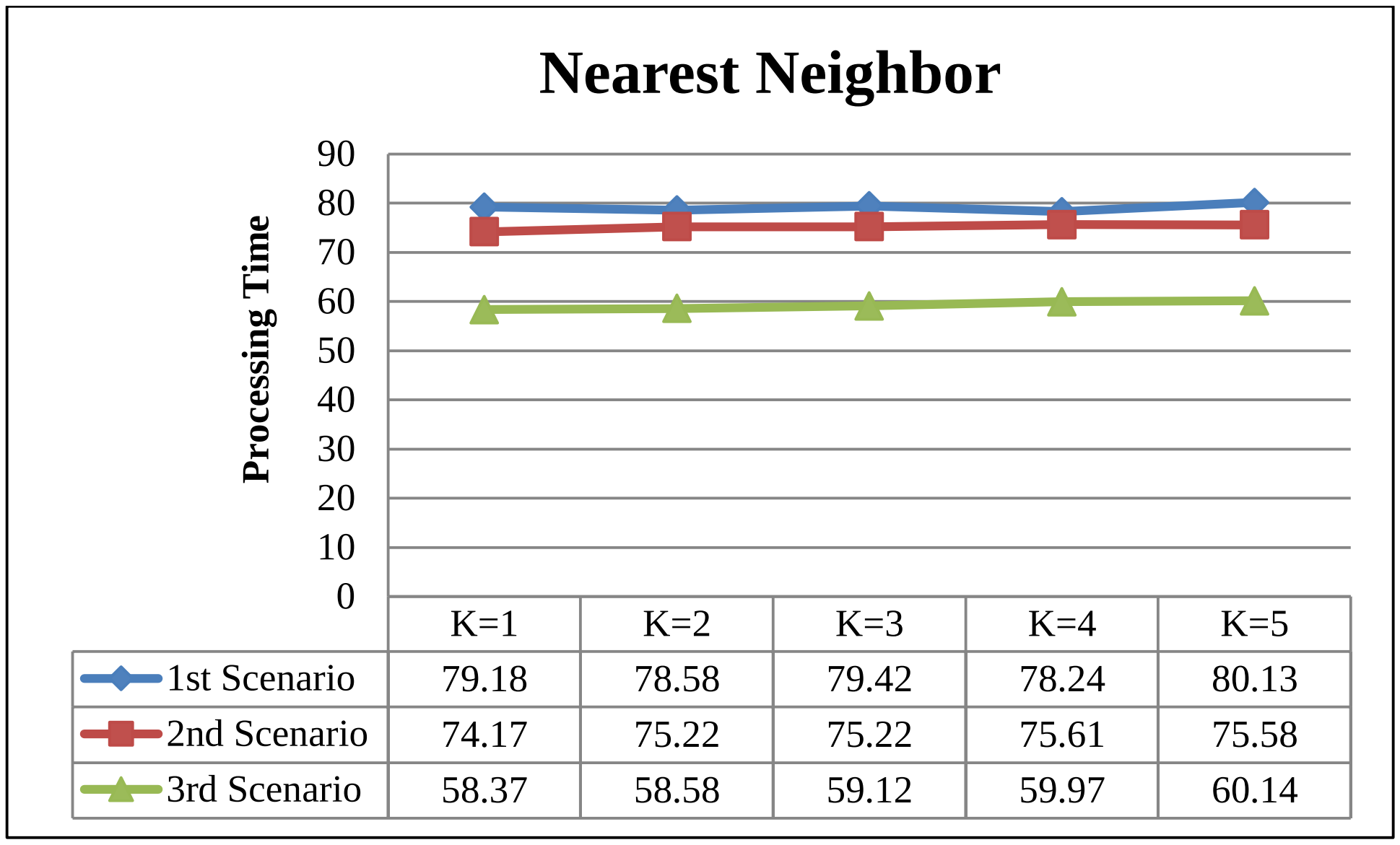

Figure 5 Processing time of nearest neighbor with different k

By using TOPSIS method, the best k for all scenarios is given in Table 4 for different assumptions.

Table 4 The best k for k-nearest neighbor method

| Assumptions Index | First Scenario | Second Scenario | Third Scenario |
|---|---|---|---|
| 1 | K=2 | K=2 | K=2 |
| 2 | K=2 | K=2 | K=2 |
| 3 | K=2 | K=2 | K=2 |
| 4 | K=3 | K=3 | K=3 |

### 3.5. Experiment results on a great range of the classifiers

Now we compare different classification methods introduced before for all three scenarios based on the functional criteria and their assigned weights. In what follows the tuned up multi-layer perceptron and k-nearest neighbor in the previous subsections are considered with respect to each scenario and every weighting assumptions.

Figure 6 presents the accuracy, sensitivity and specificity of classification methods, Figure 7 displays the processing time of classification methods. As it is shown in the figures the highest accuracy and sensitivity and specificity are obtained first for k-nearest neighbor and then for random forest. High processing time is one of the drawbacks of k-nearest neighbor method, while processing time of random forest is strongly less than the k-nearest neighbor method.

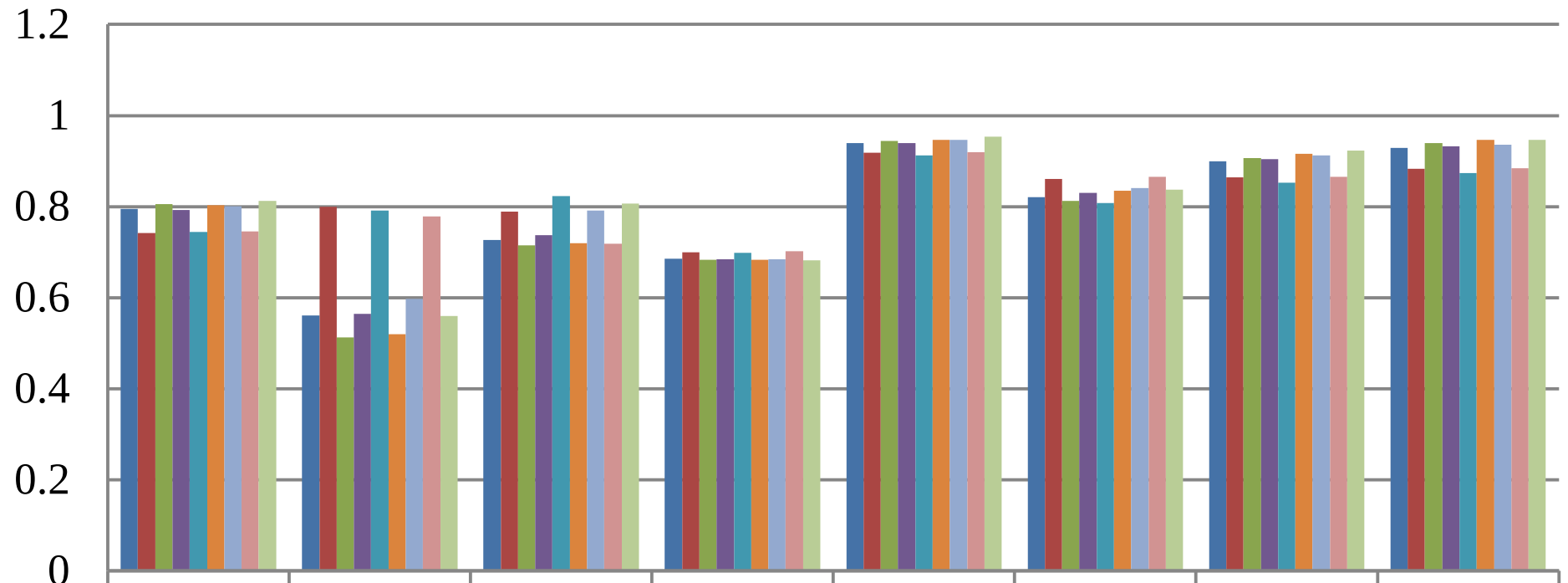

| | Bayes Network | Naïve Bayes | Multi Layer Perceptron | Sequential Minimal Optimization | K-Nearest Neighbours | Jrip | C4.5 | Random Forest |
|---|---|---|---|---|---|---|---|---|
| 1st Scenario: Accuracy | 0.795 | 0.561 | 0.727 | 0.686 | 0.939 | 0.821 | 0.899 | 0.929 |
| 1st Scenario: Sensitivity | 0.742 | 0.8 | 0.789 | 0.7 | 0.918 | 0.861 | 0.864 | 0.883 |
| 1st Scenario: Specificity | 0.806 | 0.513 | 0.715 | 0.683 | 0.944 | 0.813 | 0.907 | 0.939 |
| 2nd Scenario: Accuracy | 0.793 | 0.565 | 0.737 | 0.685 | 0.94 | 0.83 | 0.904 | 0.933 |
| 2nd Scenario: Sensitivity | 0.744 | 0.792 | 0.823 | 0.699 | 0.913 | 0.808 | 0.852 | 0.874 |
| 2nd Scenario: Specificity | 0.803 | 0.52 | 0.72 | 0.683 | 0.946 | 0.835 | 0.916 | 0.946 |
| 3rd Scenario: Accuracy | 0.801 | 0.597 | 0.792 | 0.685 | 0.947 | 0.841 | 0.913 | 0.936 |
| 3rd Scenario: Sensitivity | 0.746 | 0.779 | 0.719 | 0.702 | 0.92 | 0.865 | 0.866 | 0.884 |
| 3rd Scenario: Specificity | 0.813 | 0.56 | 0.807 | 0.682 | 0.954 | 0.837 | 0.923 | 0.947 |

Figure 6 Comparison between accuracy, sensitivity and specificity of classifiers

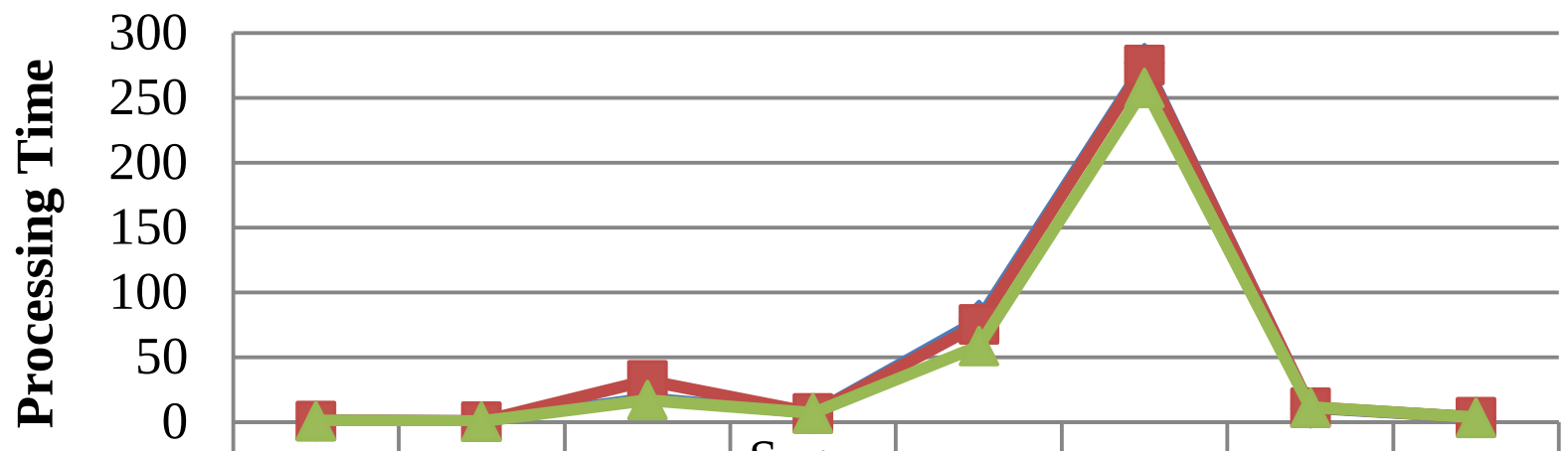

| | Bayes Network | Naïve Bayes | Multi Layer Perceptron | Sequential Minimal Optimization | K-Nearest Neighbours | Jrip | C4.5 | Random Forest |
|---|---|---|---|---|---|---|---|---|
| 1st Scenario | 0.78 | 0.34 | 18.29 | 6.85 | 78.58 | 276.7 | 10.83 | 3.68 |
| 2nd Scenario | 0.8 | 0.32 | 31.95 | 6.89 | 75.22 | 275.27 | 11.23 | 3.53 |
| 3rd Scenario | 0.84 | 0.3 | 16.96 | 6.88 | 58.58 | 257.67 | 11.32 | 3.45 |

Figure 7 Comparison between processing time of classifiers

Now using TOPSIS, it is possible to select the most effective classification method for each scenario. The values of decision matrix are given in Figure 6 and Figure 7.

Table 5 Effective classifiers for different scenarios under various assumptions

| | First Scenario | Second Scenario | Third Scenario |
|---|---|---|---|
| Assumption 1 | Random Forest | Random Forest | Random Forest |
| Assumption 2 | 2-Nearest Neighbor | 2-Nearest Neighbor | 2-Nearest Neighbor |
| Assumption 3 | Random Forest | Random Forest | Random Forest |
| Assumption 4 | Random Forest | Random Forest | Random Forest |

As it is displayed in Table 5 , using the first assumption on weights, in all three scenarios, random forest appears to be the best method for classification. Since random forest in the third scenario has highest sensitivity, specificity and accuracy, and also less processing time, random forest with 80% training data (third scenario) can be selected as the most effective classifier method to detect warning and safe situations for rear-end collisions. Its accuracy, detecting rate of warning situations and detection rate of safe situations are 93.6%, 88.4 % and 94.7 %, respectively.

The proposed random forest includes a set of decision trees in which traversing decision tree from root to leaves provides some controlling rules. In the proposed random forest, since the rules are extracted from decision tree, the parameters' thresholds and indicator’s thresholds are different and they will be initialized according to the other parameters. Some of the obtained rules are as the following:

- **If** "Speed<=73 km/h" & "DeltaX∈[4.7,8.6] m" & "DeltaV∈[-13.5,-3.2] m/s" & "TimeGap<=1.5 s " & "TimeToCollision<=2.8 s" , **Then** "Warning with frequency=(320.26, 24.5)"
- **If** "Speed∈[69,88] km/h " & "DeltaX∈[48,69] m" & "TimeGap∈[2.3,3.1] s" & "TimeToCollision>5.25 s" **Then** "Safe with frequency 66.92"

Based on the used data, the first rule demonstrates that when the speed of follower vehicle is less than 73 km/h and distance of leader vehicle from follower one is between 4.7 m and 8.6 m and relative speed of leader vehicle from follower one is between -13.5 m/s and -3.2 m/s and also the time gap is less than 1.5 s and the time to collision is less than 2.8 s, this status is warning. This rule is supported by 320.26 samples (pattern's weight) in data set where for 24.5 of samples (pattern's weight) the conclusion is not valid (not correctly predicted). In addition, the second rule shows when the speed of follower vehicle is between 69 km/h and 88 km/h, the distance of two vehicles is between 48 m and 69 m and time gap belongs to [2.3,3.1] s and the time to collision is larger than 5.25, the status is safe. This rule is supported by 66.92 samples (pattern's weights) and the results were always correctly predicted.

As it is previously mentioned, in this experiment we consider the near-crashes' data of rear-end collisions into account where the driver's reaction is braking. Now, we will consider the generalization capability of the proposed system over data of crashes and data of changing the line. Figure 8 shows the results of the generalization capability of the proposed system.

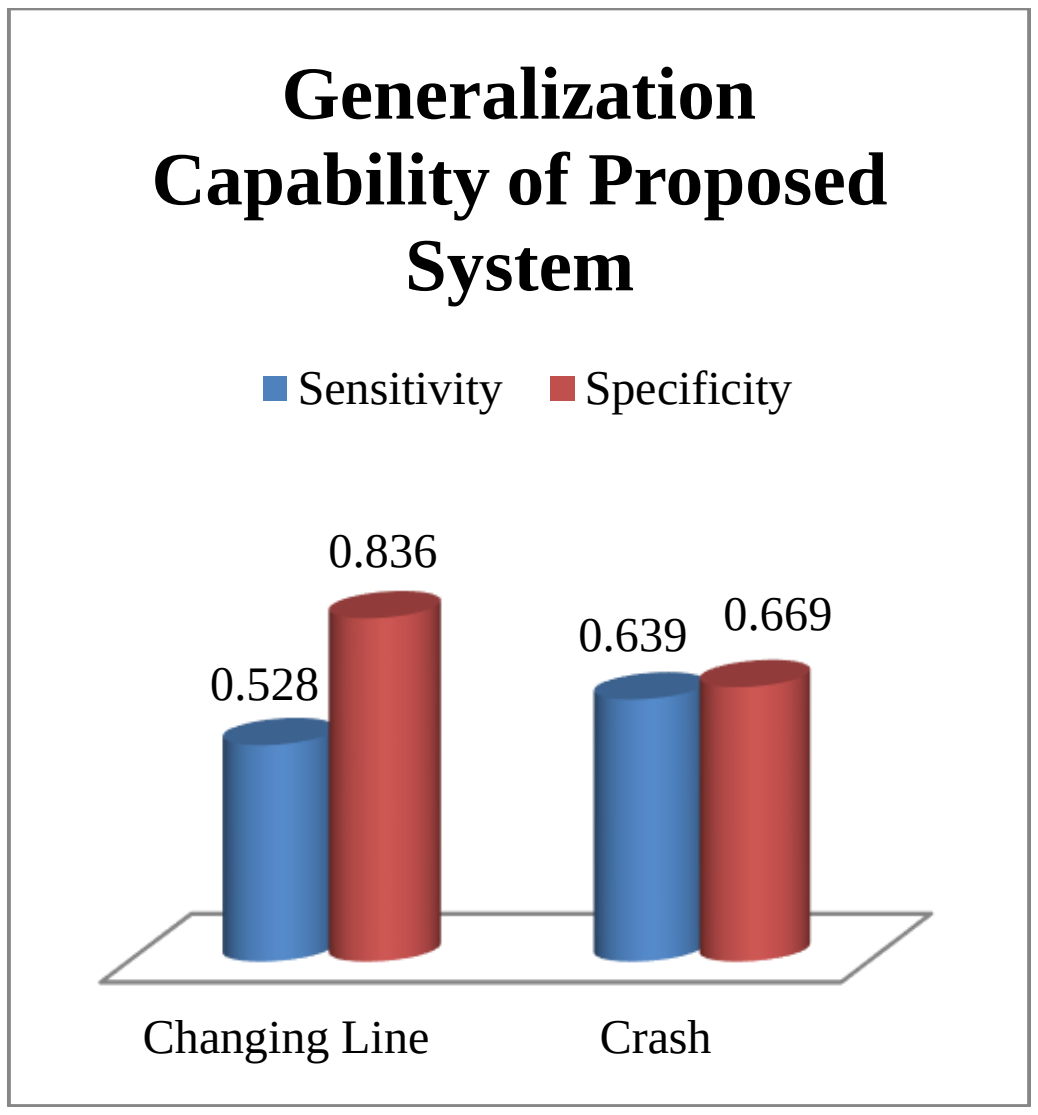

Figure 8 Generalization capability of proposed system: random forest

### 3.6. Comparison between the proposed system and perceptual-based systems

The most popular criteria which are used for evaluation of the safe and the warning situations are time to collision and time gap. Moreover time to collision is used in “Honda” and “Hirst and Graham” algorithms in the category of perceptual-based systems. If the value of each criterion of time to collision or time gap is equal or less than their threshold value, the situation is called warning, otherwise it is called safe. When the relative distance of two vehicles is less than or equal to warning distance which is calculated by “Honda” and “Hirst and Graham”algorithms, the situation is warning, otherwise it is detected as safe.

The success of perceptual-based warning systems depends on the appropriate and correct selection of threshold values. Due to the critical threshold of time to collision, there are some research studies to suggest values which are gathered inTable 6 .

Table 6  Previous results about recommended time-to-collision

| Recommended time-to-collision | Source |
|---|---|
| 2 | [23] |
| 2.2 | [16] |
| 3 | [17] |
| 3.5 | [49] |
| 4 | [50] |

In addition to the previously suggested threshold values, we will calculate the threshold value for the available vehicles trajectory data. To do this, we used the pruned decision tree C4.5classifier using cost sensitive learning.

Now, we are looking for a threshold value for vehicles trajectory data used in this study. So we used the property "time to collision" for data samples and pruned C4.5 decision tree as classifier and then classification is applied on the data. The results of the obtained decision tree using the cost sensitive learning is shown in Figure 9. Note that 0 presents the safe situation and 1 refers to the warning situation. Based on the Figure 9, using cost-sensitive learning, the threshold value is 6.5. It is also worth to mention that the threshold value of the time to collision criterion means the warning situation for the values between 0 to threshold value, otherwise it means safe situation.

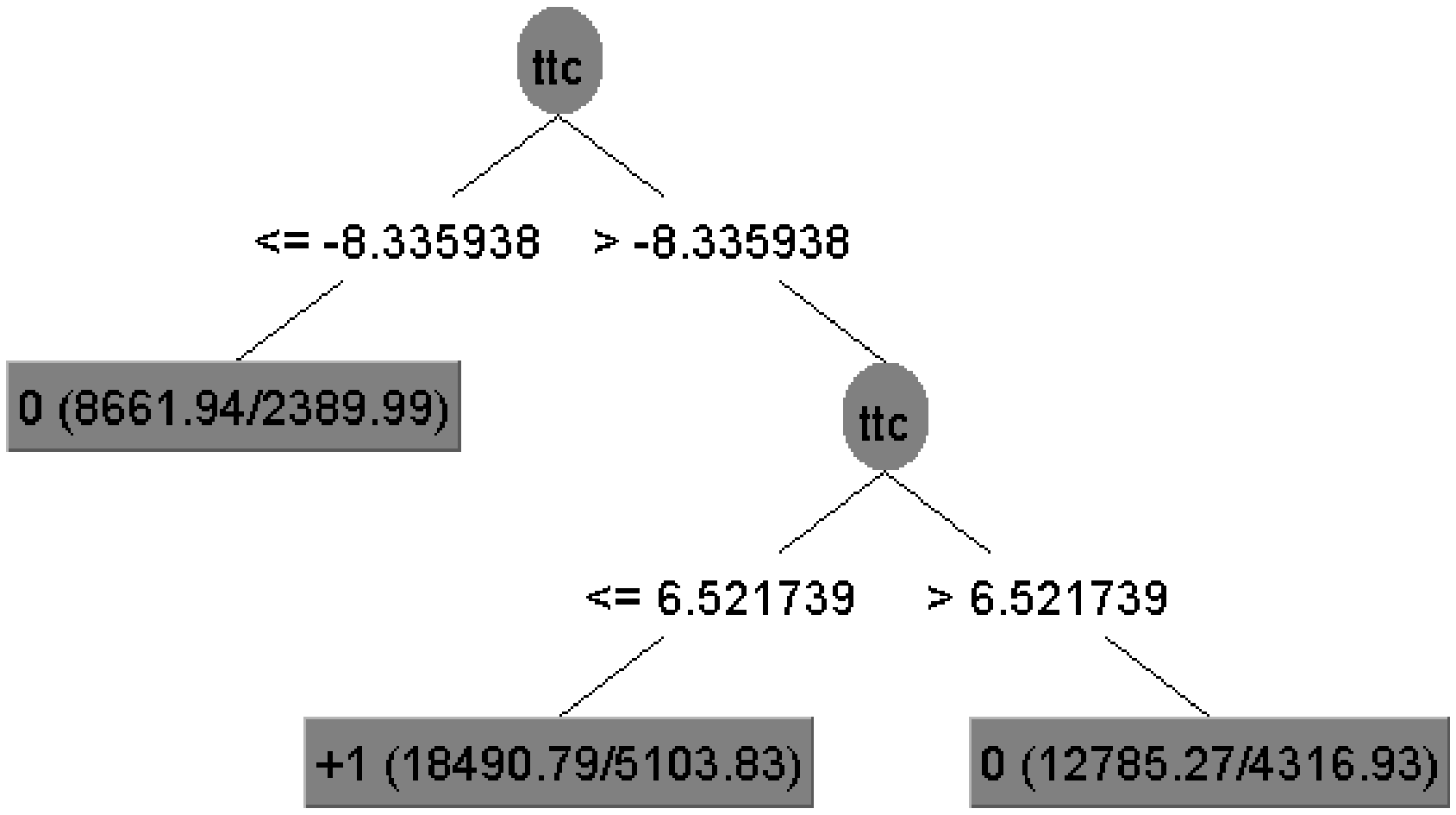

Figure 9 Critical threshold for time to collision, using cost sensitive learning

Figure 10 shows the sensitivity and specificity for the criterion of time to collision with different threshold values, in addition it shows sensitivity and specificity for two algorithms “Honda” and “Hirst and Graham” taking different threshold values for the criterion time to collision into account.

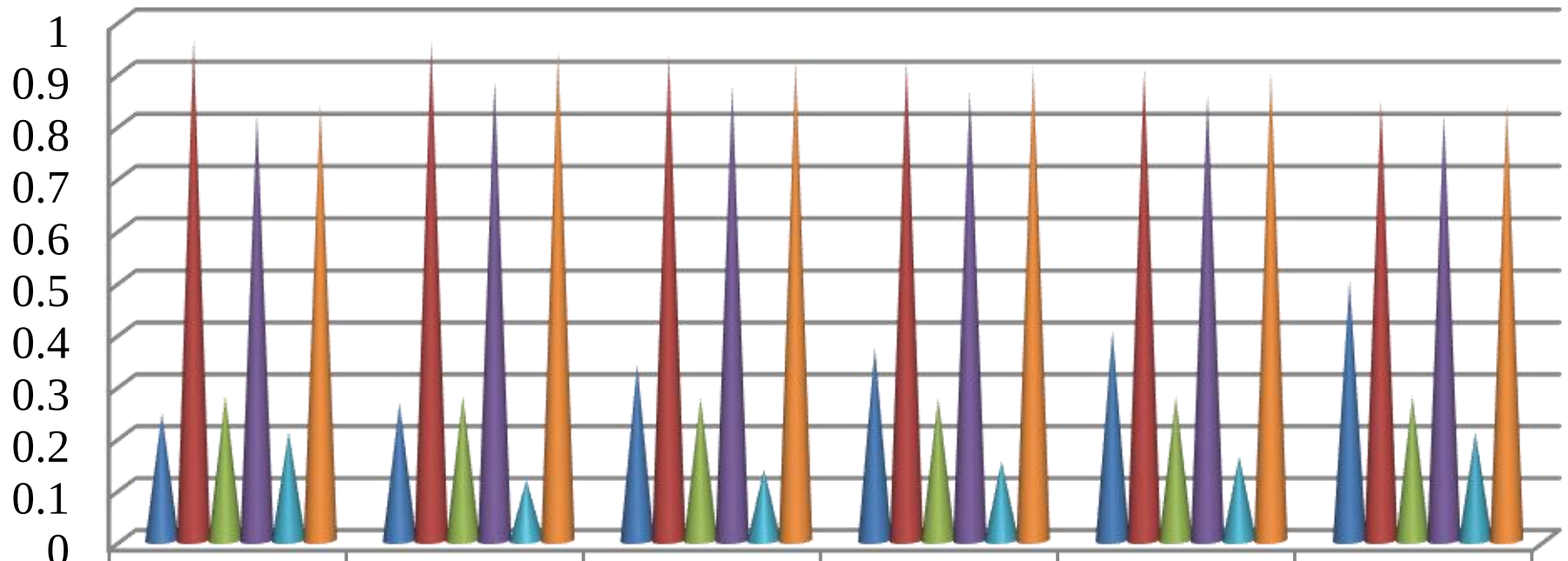

| | 2 | 2.2 | 3 | 3.5 | 4 | 6.5 |
|---|---|---|---|---|---|---|
| TTC Sensitivity | 0.244 | 0.265 | 0.339 | 0.372 | 0.404 | 0.501 |
| TTC Specificity | 0.962 | 0.957 | 0.933 | 0.92 | 0.905 | 0.846 |
| Honda Sensitivity | 0.279 | 0.279 | 0.275 | 0.275 | 0.277 | 0.279 |
| Honda Specificity | 0.819 | 0.885 | 0.874 | 0.865 | 0.857 | 0.819 |
| Hirst&Graham Sensitivity | 0.209 | 0.117 | 0.137 | 0.153 | 0.162 | 0.209 |
| Hirst&Graham Specificity | 0.84 | 0.941 | 0.922 | 0.911 | 0.899 | 0.84 |

Figure 10 Sensitivity and specificity for time to collision, Honda and Hirst&Graham with different critical thresholds

The experimental results show that the systems which work based on time to collision criterion are more precise for detecting safe situations but they are weak in detecting warning situations. So the ratio of warning situations detected by these systems is smaller than the other systems. This observation is referred in the article [51].

About the critical threshold of time gap there are some research studies where the suggested values are shown on Table 7 .

Table 7 Previous results about recommended time-gap

| Recommended time-gap | Source |
|---|---|
| 1.6 s or more (no secondary task distraction)<br>2.08 s or more (being distracted by secondary tasks) | [52] |
| 1.5-2.49 s (motorway)<br>1.66-3.21 s (rural wary) | [53] |
| 2 s or more | [54] |
| 1.1 s (young)<br>1.5 s (middle aged)<br>2.1 s (older) | [55] |
| 1.1-1.8 s | [56] |

Now we will find out the critical threshold value for the vehicles trajectory data. To do this, we selected time gap of sample data and pruned C4.5 decision tree as classifier, then we classify the data. The decision tree obtained from cost-sensitive learning is shown in Figure 11. Please note that 0 means safe situations and 1 represents warming situations.

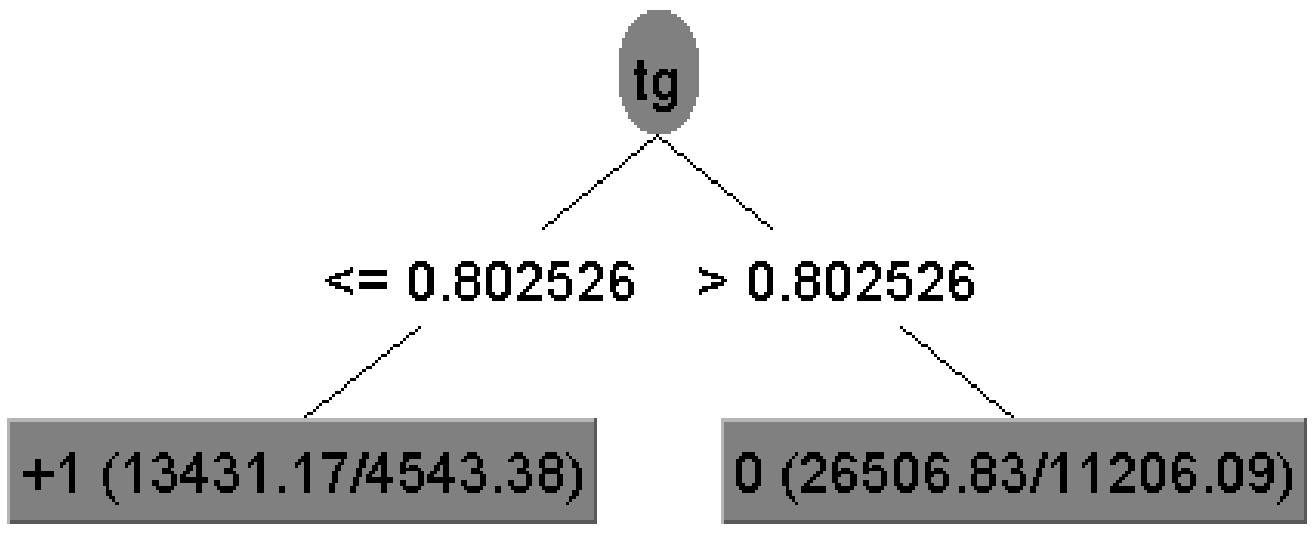

Figure 11 Critical threshold for time gap, using cost sensitive learning

As Figure 11 shows, the critical threshold of time gap is selected as 0.8 according to the vehicles trajectory data. Figure 12 shows sensitivity and specificity for time gap criterion with different threshold values.

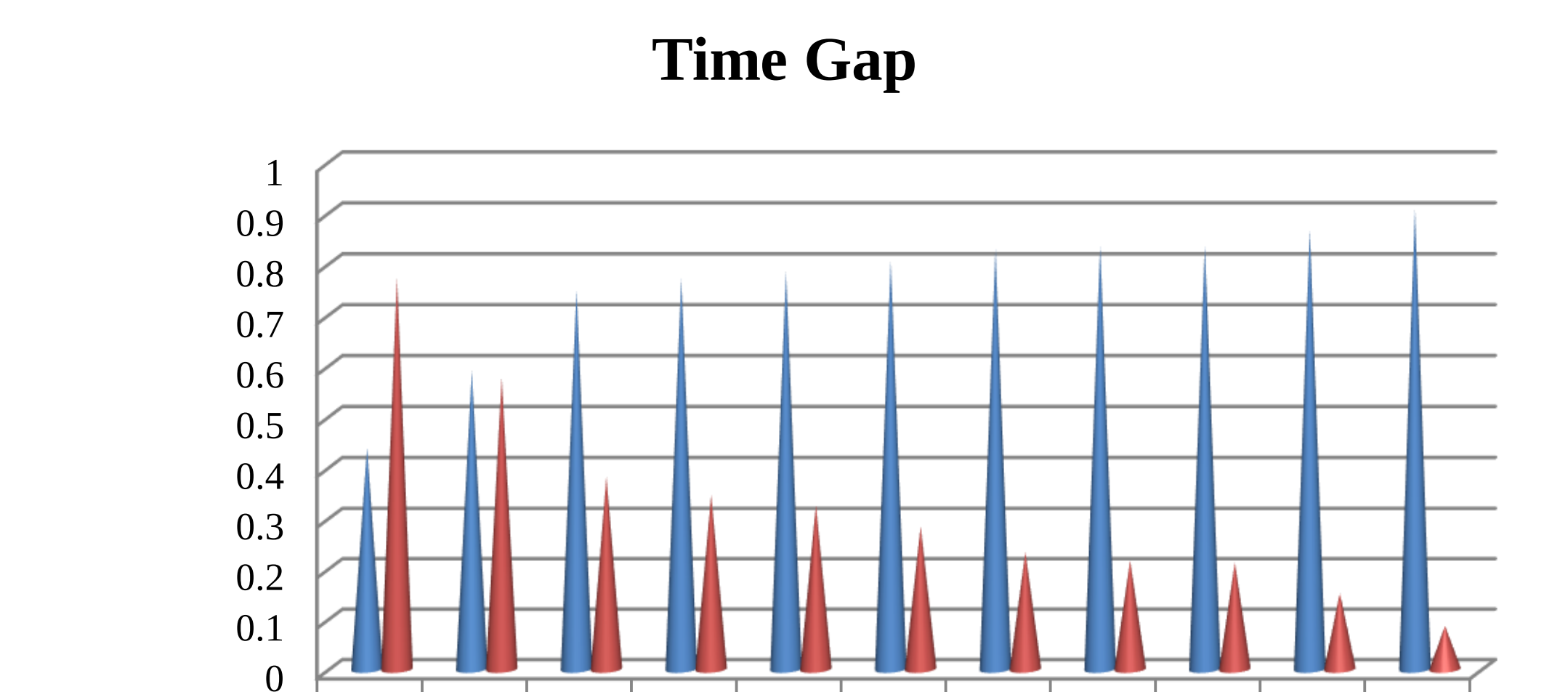

| | 0.8 | 1.1 | 1.5 | 1.6 | 1.66 | 1.8 | 2 | 2.08 | 2.1 | 2.49 | 3.21 |
|---|---|---|---|---|---|---|---|---|---|---|---|
| Sensitivity | 0.439 | 0.589 | 0.748 | 0.771 | 0.784 | 0.805 | 0.829 | 0.835 | 0.836 | 0.867 | 0.907 |
| Specificity | 0.773 | 0.576 | 0.382 | 0.346 | 0.326 | 0.284 | 0.233 | 0.216 | 0.212 | 0.151 | 0.086 |

Figure 12 Sensitivity and specificity for time gap with different critical thresholds

As Figure 12 shows, the systems which work based on time gap criterion with threshold value larger than 1.5 are more precise for detecting warning situations but they are weak in detecting safe situations. As a result, the percentage of situations detected as warning by these systems is fairly more than other systems. The results are referred in [51].

The Figure 13 shows the results of comparing the proposed classifier system, the random forest, with the perceptual based systems, regarding the threshold values found from the vehicles trajectory data.

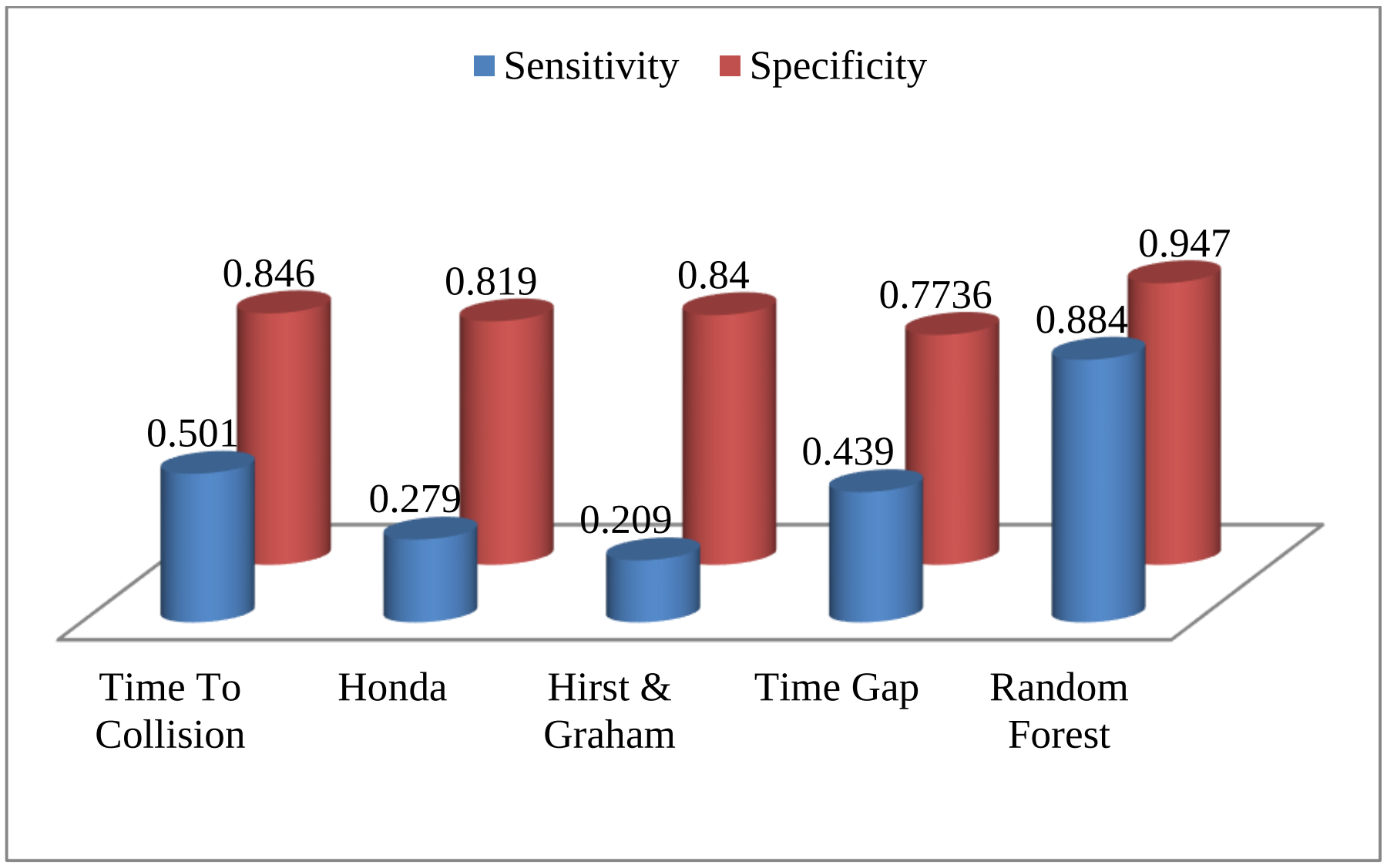

Figure 13 Comparison between proposed system and perceptual-based systems

As Figure 13 shows, the specificity and sensitivity of the proposed random forest is different from the other perceptual-based algorithms. We can say the perceptual based algorithms are using only one criterion and one constant threshold value for the criterion therefore they cannot have good precision.

### 3.7. Comparison between proposed system and kinematic-based systems

In kinematic-based systems, the safe distance for the follower vehicle is the criterion for issuing warning; based on that distance, warning will be issued. The most important algorithms for calculating safe distance of follower vehicle include MAZDA, stop distance and PATH. In vehicles trajectory data that we used in this study, first the distance between vehicles are calculated and if this distance is equal or less than warning distance calculated by each one of the algorithms Mazda, stop distance and PATH, we called the situation warning, otherwise it is safe and there is no need to issue warning based on each algorithm. Comparing warning and safe situation, detected by each algorithm with real situation for each data sample, we could obtain sensitivity and specificity for each algorithm based on vehicles trajectory data. Figure 14 shows the results of the proposed classifier system aka random forest and kinematic-based systems.

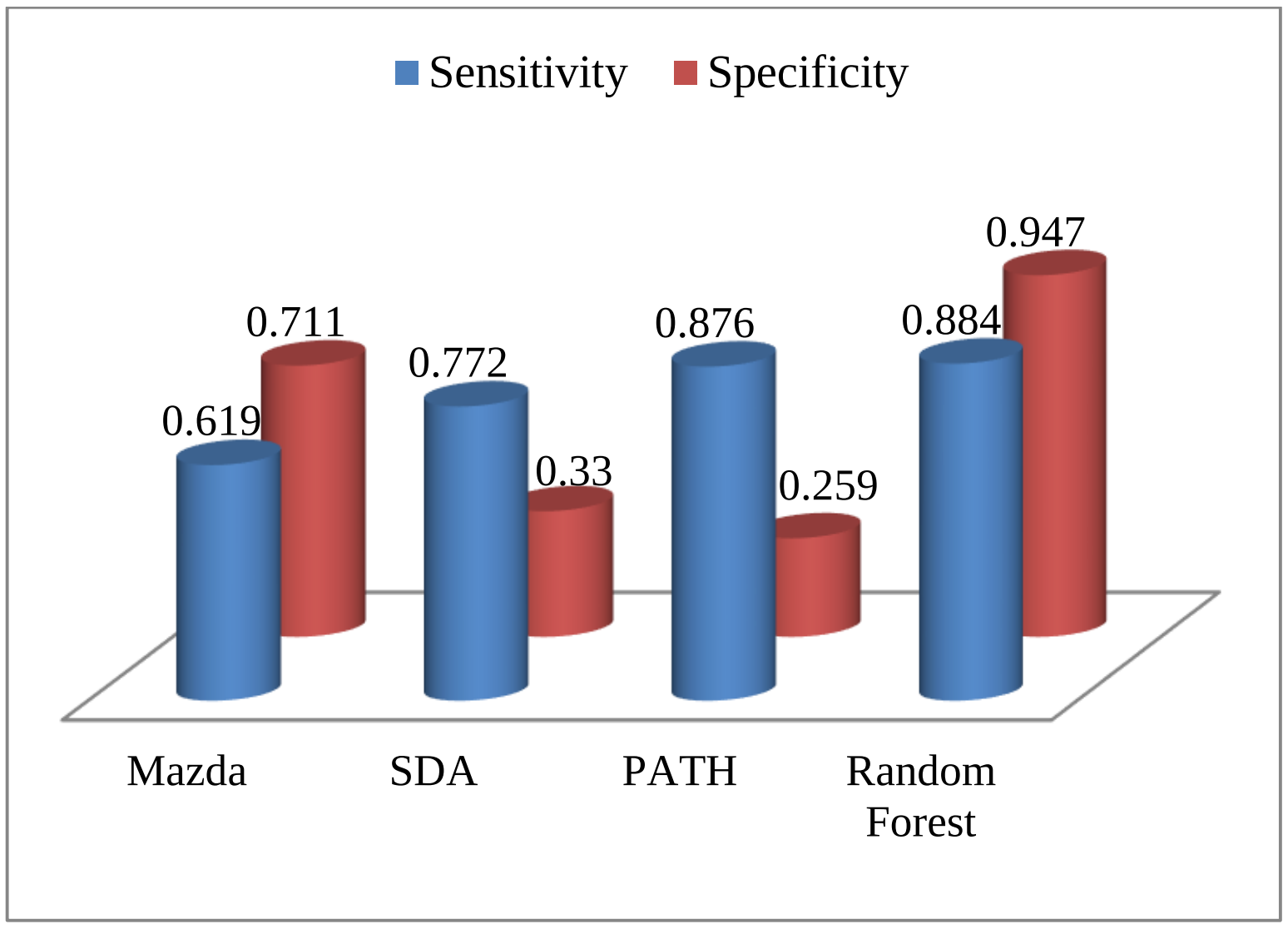

Figure 14 Comparison between proposed system and kinematic-based systems

As Figure 14 shows, there is a huge difference between the proposed system and all the other kinematic-based algorithms in view of specificity and sensitivity. Actually, kinematic-based algorithms will not result in desirable precision because they used constant and predefined values for two parameters 1) reaction time of follower vehicle and 2) maximum decreasing rates of vehicle's speeds in addition to the assumption of fixed decreasing rate of follower vehicle's speed. Bunch of researches have focused on finding appropriate reaction time of driver and maximum decreasing rate of vehicle's speed, where different values are taken into account for these two parameters. Therefore, low precision of these algorithms is a result of pre-defined and constant values.

In the algorithm stop-distance for different values of constant parameters, we have Table 8 . In the first scenario, the same pre-defined parameters used in the algorithm stop-distance are used, in the second scenario we used the same parameters as parameters in algorithm MAZDA and in the third scenario, we used the same parameters as algorithm PATH and in the fourth scenario we used our proffered parameters.

Table 8 Initializing parameters of stop-distance algorithm

| | $a_f(m/s^2)$ | $a_l(m/s^2)$ | $\tau_{driver}(s)$ |
|---|---|---|---|
| First Scenario | 5 | 5 | 1.5 |
| Second Scenario | 6 | 8 | 0.1 |
| Third Scenario | 6 | 6 | 0.5 |
| Fourth Scenario | 7 | 7 | 0.8 |

The results of algorithm, stop-distance in view of sensitivity and specificity for parameters in Table 8 are shown in Figure 15.

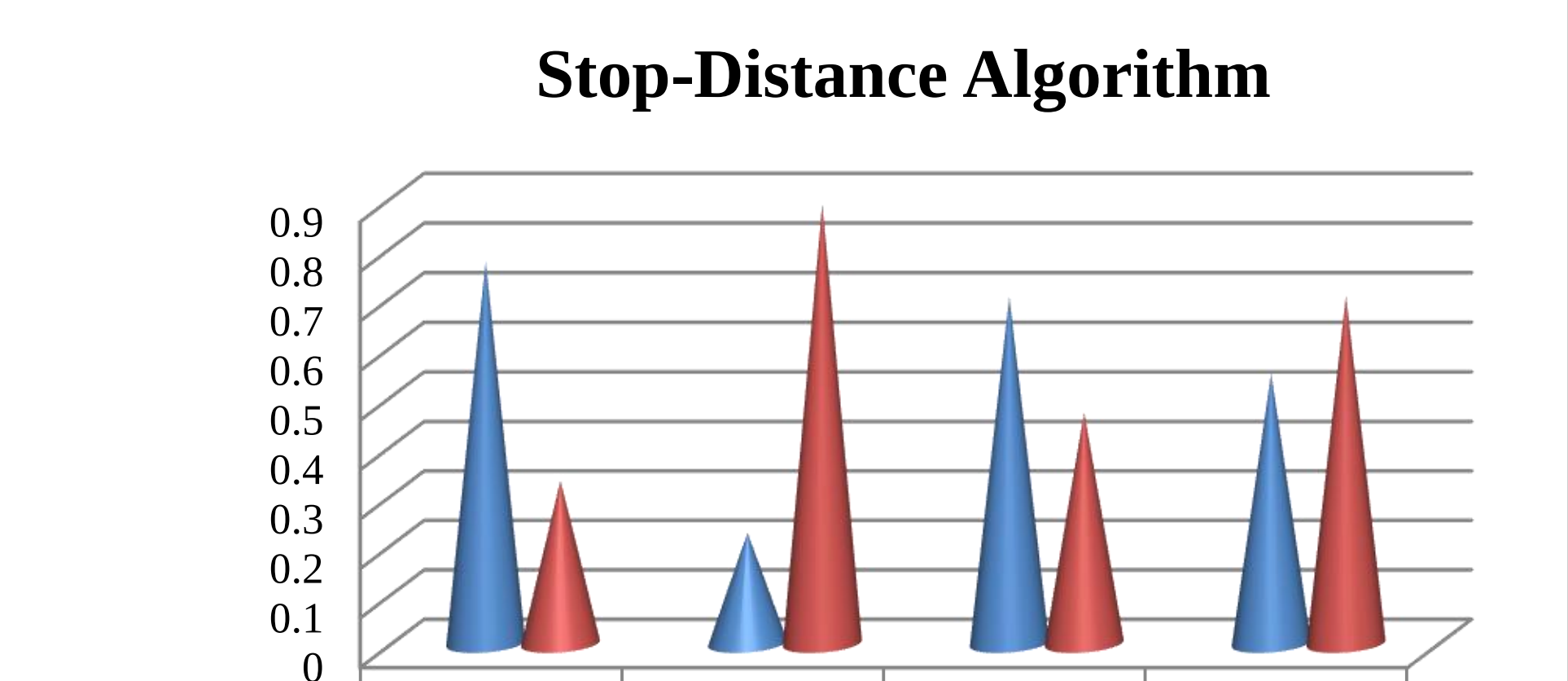

| | First Scenario | Second Scenario | Third Scenario | Fourth Scenario |
|---|---|---|---|---|
| Sensitivity | 0.772 | 0.224 | 0.701 | 0.546 |
| Specificity | 0.33 | 0.888 | 0.468 | 0.703 |

Figure 15 Stop-distance algorithm with different parameters' initializing

As Figure 15 shows, change in constant parameters in the algorithm, results in remarkable change in sensitivity and specificity. Therefore, kinematic-based systems are not a good base for warning in rear-end collisions.

### 3.8. Comparison between proposed system and existing systems - conclusion

The Figure 16 shows comparison between the proposed classifier system with kinematic-based systems and perceptual-based systems. The threshold values for perceptual-based systems are equal to the threshold value obtained from the vehicles trajectory data used in this study. As we mentioned already, the threshold value obtained from the vehicles trajectory data for the time to collision criterion is equal to 6.5 and for the time gap criterion is 0.8.

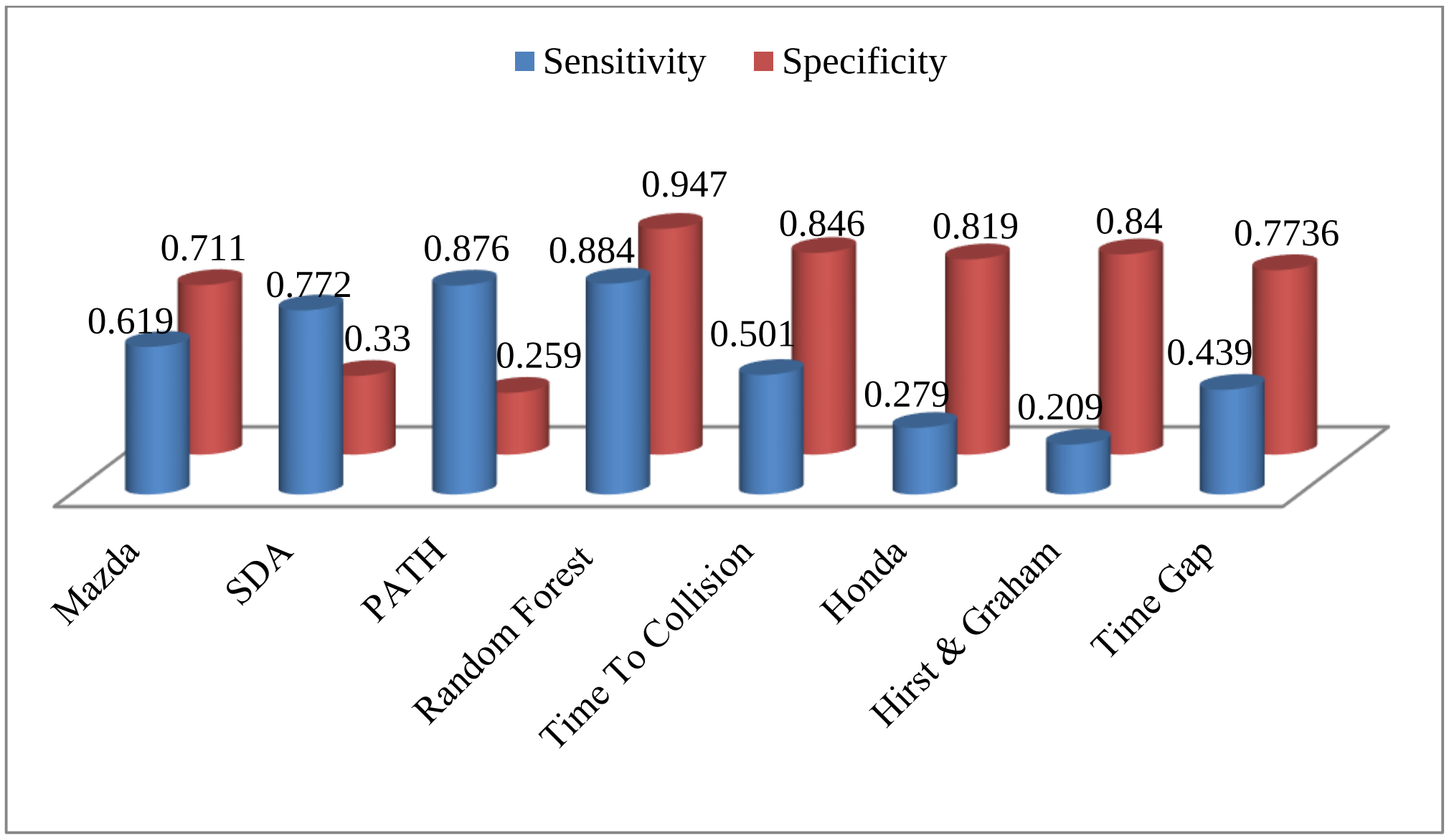

Figure 16 Comparison between the proposed classifier system with kinematic-based systems and perceptual-based systems

As we already mentioned, the kinematic-based algorithms use the constant values and predefined values for the parameters and the perceptual-based algorithms use constant threshold values, therefore both cannot result in good precision. Using only one criterion in rear-end collision warning systems is not a good approach. Time to distance criterion is weak to detect warning situations and also time gap criterion bothers the driver with frequent warning alarms.

The proposed random forest system outperforms the other systems for detecting the warning situations and safe ones with high precision for some reasons as follows: 1) using those two criteria as well as speed, relative speed, relative distance, 2) dynamic threshold values and non-constant threshold value, 3) the warning and safe situation learning is based on the naturalistic data of collisions, before event, event and after event.

## 4. Conclusions

In this paper, rear-end collision warning system based on data mining is proposed. Using the classification algorithms of Bayesian network, Naïve Bayes, MLP neural network, support vector machine, nearest neighbor, rule-based methods, decision tree and random forest; situations were classified into warning and safe classes. Since the data of two classes were imbalanced, we used the combination of cost sensitive learning and classification methods. Using sensitivity, specificity and processing time as criteria and using TOPSIS method as a multi-criteria decision making method, we obtained random forest as optimum classifier. This classifier is proposed as a rear-end collision warning system which is able to detect warning situations with the probability of 88.4% and able to detect safe situations with probability of 94.7%.The proposed rear-end collision warning classifier is compared against current algorithms which are perceptual-based and kinematic-based, to show how the proposed one outperforms all of them.

**Acknowledgment**